\def\isarxiv{1}

\def\paperTitle{T2VWorldBench: A Benchmark for Evaluating World Knowledge in Text-to-Video Generation
}

\def\paperAuthor{
Yubin Chen\thanks{\texttt{abinzzz1227@gmail.com}. San Jose State University.}
\and
Xuyang Guo\thanks{\texttt{gxy1907362699@gmail.com}. Guilin University of Electronic Technology.}
\and
Zhenmei Shi\thanks{\texttt{zhmeishi@cs.wisc.edu}. University of Wisconsin-Madison.}
\and
Zhao Song\thanks{\texttt{magic.linuxkde@gmail.com}. University of California, Berkeley.}
\and 
Jiahao Zhang\thanks{\texttt{ml.jiahaozhang02@gmail.com}.}
}

\ifdefined\isarxiv
\ifdefined\paperID
\PackageError{CaoChongChengXiang}{You must remove paperID for arXiv submission!}
\fi
\documentclass[11pt]{article}
\usepackage[numbers]{natbib}
\else
\documentclass[10pt,twocolumn,letterpaper]{article}

\usepackage[review,algorithms]{wacv}      
\fi

\ifdefined\isarxiv

\usepackage{amsmath}
\usepackage{amsthm}
\usepackage{amssymb}
\usepackage{algorithm}
\usepackage{subfig}
\usepackage{algpseudocode}
\usepackage{graphicx}
\usepackage{grffile}
\usepackage{wrapfig,epsfig}
\usepackage{url}
\usepackage{xcolor}
\usepackage{epstopdf}

\usepackage{bbm}
\usepackage{dsfont}

\else 
\usepackage{amsmath}
\usepackage{amsthm}
\usepackage{amssymb}
\usepackage{algorithm}
\usepackage{algpseudocode}
\usepackage{graphicx}
\usepackage{grffile}
\usepackage{wrapfig,epsfig}
\usepackage{url}
\usepackage{xcolor}
\usepackage{epstopdf}

\usepackage{bbm}
\usepackage{dsfont}

\usepackage{booktabs} 

\definecolor{wacvblue}{rgb}{0.21,0.49,0.74}
\usepackage[pagebackref,breaklinks,colorlinks,allcolors=wacvblue]{hyperref}

\def\wacvPaperID{\paperID} 
\def\confName{WACV}
\def\confYear{2026}

\fi
 
\allowdisplaybreaks

\ifdefined\isarxiv

\usepackage{tikz}
\usepackage{hyperref}  
\hypersetup{colorlinks=true,citecolor=blue,linkcolor=blue} 
\usetikzlibrary{arrows}
\usepackage[margin=1in]{geometry}

\else

\fi
 
\graphicspath{{./figs/}}

\theoremstyle{plain}
\newtheorem{theorem}{Theorem}[section]

\newtheorem{observation}[theorem]{Observation}

\usepackage{booktabs} 

\begin{document}

\ifdefined\isarxiv

\date{}
\title{\paperTitle}
\author{\paperAuthor}

\else

\title{\paperTitle}

\author{First Author\\
Institution1\\
Institution1 address\\
{\tt\small firstauthor@i1.org}
\and
Second Author\\
Institution2\\
First line of institution2 address\\
{\tt\small secondauthor@i2.org}
}
\maketitle
\fi

\ifdefined\isarxiv
\begin{titlepage}
  \maketitle
  \begin{abstract}
    Text-to-video (T2V) models have shown remarkable performance in generating visually reasonable scenes, while their capability to leverage world knowledge for ensuring semantic consistency and factual accuracy remains largely understudied. In response to this challenge, we propose {\bf T2VWorldBench}, the first systematic evaluation framework for evaluating the world knowledge generation abilities of text-to-video models, covering 6 major categories, 60 subcategories, and 1,200 prompts across a wide range of domains, including physics, nature, activity, culture, causality, and object. To address both human preference and scalable evaluation, our benchmark incorporates both human evaluation and automated evaluation using vision-language models (VLMs). We evaluated the 10 most advanced text-to-video models currently available, ranging from open source to commercial models, and found that most models are unable to understand world knowledge and generate truly correct videos. These findings point out a critical gap in the capability of current text-to-video models to leverage world knowledge, providing valuable research opportunities and entry points for constructing models with robust capabilities for commonsense reasoning and factual generation.

  \end{abstract}
  \thispagestyle{empty}
\end{titlepage}


\else

\begin{abstract}
Text-to-video (T2V) models have shown remarkable performance in generating visually reasonable scenes, while their capability to leverage world knowledge for ensuring semantic consistency and factual accuracy remains largely understudied. In response to this challenge, we propose {\bf T2VWorldBench}, the first systematic evaluation framework for evaluating the world knowledge generation abilities of text-to-video models, covering 6 major categories, 60 subcategories, and 1,200 prompts across a wide range of domains, including physics, nature, activity, culture, causality, and object. To address both human preference and scalable evaluation, our benchmark incorporates both human evaluation and automated evaluation using vision-language models (VLMs). We evaluated the 10 most advanced text-to-video models currently available, ranging from open source to commercial models, and found that most models are unable to understand world knowledge and generate truly correct videos. These findings point out a critical gap in the capability of current text-to-video models to leverage world knowledge, providing valuable research opportunities and entry points for constructing models with robust capabilities for commonsense reasoning and factual generation.
\end{abstract}

\fi



\section{Introduction}\label{sec:intro}

Recent progress in generative models has greatly improved the performance of text-to-video (T2V) models in several aspects~\cite {zzzk23,brl+23,hdz+23,zwl+24}, including video editing~\cite{chm23,yzfy25}, motion consistency~\cite{hwc+25,wls+25}, and object consistency~\cite{shl+25,xyyg25}, which promotes exploration of video generation~\cite{sph+23,yhx+24,wly+25}. Text-to-video models do not generate static images, but they model a real physical world to create highly aesthetic and realistic videos. Currently, advanced T2V models such as Sora~\cite{sora} and Kling~\cite{kling} are able to generate realistic videos that conform to the laws of physics based on user prompts. These amazing video generation technologies have significantly changed the way we interact with videos, allowing even amateurs to create cinema-level scenes with director-like precision, and receiving widespread attention from both the public and the research community. 

\begin{figure}[!ht]
    \centering
    \includegraphics[width=0.65\linewidth]{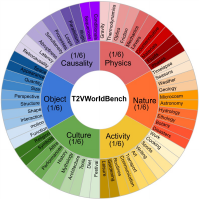}
    \caption{\textbf{Prompt Domain Distribution of T2VWorldBench}}
    \label{fig:benchmark_detail}
\end{figure}

Despite text-to-video models having achieved breakthroughs in semantic understanding and video quality~\cite{hhy+24,lcl+24,zhl+25}, a critical limitation persists: most current text-to-video models perform well under fictional prompts and fail to explore models' capacity to produce videos according to world knowledge. While recent research has investigated the capacity of T2I models to produce images based on world knowledge~\cite{zjx+25,nnz+25}, the relevant capability in the text-to-video domain remains less explored. Truly intelligent video generation requires a deep understanding of real-world phenomena, causal relationships, and commonsense reasoning, rather than just pixel manipulation~\cite{svc+24,ywp+24}. For instance, when generating a video of seed germination for educational purposes, it is important to visually depict a continuous process, where the radicle breaks through the seed coat and grows upward until green leaves eventually sprout, rather than showing a seed suddenly transforming into a seedling. Consequently, there is a growing need for dedicated research and comprehensive evaluation to test T2V models' overall understanding of world knowledge and ability to reason based on that knowledge.

To this end, we present \textbf{T2VWorldBench}, a comprehensive benchmark aimed at assessing the world knowledge capabilities of T2V models. The benchmark includes 1,200 textual prompts from six main categories (see Figure~\ref{fig:benchmark_detail} for the full taxonomy). These prompts are used to evaluate 10 state-of-the-art T2V models, covering both commercial and open-source systems, and reflecting the latest advancements in text-to-video generation as of 2025. 
To balance scalable evaluation with human preference, we adopt a mixed evaluation protocol in which both automated and human evaluations are conducted on the same four criteria: video quality, video realism, video relevance, and video consistency. For automatic evaluation, human annotators first provide detailed explanations grounded in real-world commonsense knowledge for each prompt. We then assess whether the generated videos align with these expectations using vision-language models (VLMs)~\cite{lbpl19, ldf+20, zyx+23}, enabling a trustworthy and knowledge-aware evaluation that goes beyond simple quality metrics. 
For human evaluation, multiple annotators independently review each generated video frame by frame and assign scores based on the same four criteria. 
Our key contributions are detailed below: 

\begin{itemize}
    \item To the best of our understanding, we take the initial step to introduce a text-to-video benchmark based on world knowledge, featuring six categories with 1200 prompts, comprising physics, nature, activity, culture, causality, and object.
    \item Through visual model evaluation and manual evaluation, we assess text-to-video models from 4 aspects: video quality, video realism, video relevance, and video consistency. We found that the current text-to-video models perform poorly in video generation based on world knowledge, with overall scores generally lower than 0.70. 
\end{itemize}

{\bf Roadmap.} We systematically review the relevant works of this benchmark in Section~\ref{sec:rel_works}. Section~\ref{sec:bench} presents the explicit description of the T2VWorldBench benchmark. We report the main assessment results of our evaluation framework in Section~\ref{sec:experiments}. Section~\ref{sec:conclusion} presents several concluding remarks for this paper.

\begin{table}[!ht]
    \centering
     \resizebox{0.8\linewidth}{!}{ 
     \begin{tabular}{|c|c|c|c|c|}
        \hline
        \textbf{Model Name} & \textbf{Year} & \textbf{Organization} & \textbf{\# Params} & \textbf{Open} \\
        \hline
        Sora~\cite{sora} & 2024 & OpenAI & N/A & No \\
        \hline
        Mochi-1~\cite{mochi1} & 2024 & Genmo & 10B & Yes \\
        \hline
        PixVerse V4.5~\cite{pixVerse} & 2024 & AISphere & N/A & No \\
        \hline
        Kling~\cite{kling} & 2024 & Kuai & N/A & No \\
        \hline
        Dreamina~\cite{dreamina} & 2024 & ByteDance & N/A & No \\
        \hline
        Qingying~\cite{qingying} & 2024 & Zhipu & 5B & Yes \\
        \hline
        LTX Video~\cite{hcb+24} & 2024 & Lightricks & 2B & Yes  \\
        \hline
        Pika 2.2~\cite{pika2.2} & 2025 & Pika Labs & N/A & No \\
        \hline
        Hailuo~\cite{hailuo} & 2025 & MiniMax & N/A & No \\
        \hline
        Wan 2.1~\cite{wan2.1} & 2025 & Alibaba & 14B & Yes \\
        \hline
    \end{tabular}
    }
    \caption{\textbf{Overview of 10 Evaluated Text-to-Video Models in Our Benchmark.}}
    \label{tab:models}
\end{table}

\section{Related Work}\label{sec:rel_works}

{\bf Text-to-video Generation.} Recently, on the basis of the success of text-to-image models, diffusion models have achieved significant progress in text-to-video (T2V) generation~\cite{brl+23, cxh+23, lcz+23, sph+23}. Early work on text-to-video focused primarily on GAN~\cite{gpm+14, rmc16, kla19} and VAE~\cite{kw14, rmw14, hmp+17}, limited by the model's generalization and semantic understanding capabilities. Nowadays, through training with large-scale data, T2V models could generate realistic and visually appealing videos~\cite{ywl+23, wy24, ojk+24, nxz+25}, such as Sora~\cite{sora} through its similar diffusion transform architecture, which integrates the generative power of diffusion models with the sequence modeling ability of transform, generating realistic videos with spatial and temporal sequences conforming to human aesthetics based on large-scale pre-training. Similarly, Kling~\cite{kling} integrates physical modeling, controllable camera systems, and efficient diffusion architecture to enable the model to generate convincing and clear videos while ensuring semantic coherence. These T2V models demonstrate impressive ability in generating videos with high visual quality, semantic consistency, and scene diversity~\cite{gzh+23, ytz+24}. However, current T2V models exhibit limitations in incorporating world knowledge into video generation~\cite{sph+23, cxl+24, cwl+24}, which serves as one of our main motivations.

{\bf Text-to-video Evaluation.} As the development of T2V models is getting faster, it is increasingly important to test the performance of the T2V models in all areas~\cite{lyz+24,gls+25,cgl+25_rich,lss+25_gm}. 
This allows us to explore the fundamental limitations of such generative models~\cite{gkl+25,kll+25,cll+25_var,hwsl24}, and point out many future directions like high-order flow matching~\cite{lss+25_hofar,lss+25_nrflow,cgl+25_homo,gll+25}, lazy propagation~\cite{ssz+25_dit,nwz+24}, and theoretical guarantee~\cite{csy25_vlfm,lsy25}. 
Initially, using Inception Score (IS)~\cite{sgz+16}, Fr\'{e}chet inception distance (FID)~\cite{hru+18}, and Fr\'{e}chet Video Distance (FVD)~\cite{uvk+19} as metrics to evaluate the video quality. For semantic consistency, CLIPScore~\cite{hhf+21} is introduced as a metric to evaluate the similarity of the text prompt and generated video by leveraging the CLIP model~\cite{rkh+21}. While early assessment metrics performed well in low-level perception and static semantic alignment, they still face challenges in capturing temporal coherence, physical modeling, and fine-grained explanation. Several new benchmarks have been proposed to improve T2V evaluation, among which are Comprehensive assessment~\cite{hhy+24, hzx+24}, numerical constraints~\cite{ghh+25, cgh+25}, dynamic consistency~\cite{lyz+24, jxth24}, fine-grained assessment~\cite{llr+23, gls+25}, combination of multiple properties~\cite{fls+24, shl+25}, physical principle constraints~\cite{msl+24, ghs+25}. To be specific, \cite{jxth24} introduces a temporal dynamics benchmark that conducts a hierarchical evaluation of 16 key temporal dimensions, including multiple evaluation metrics such as CLIPScore, BLIPScore, and VQA Score. \cite{hhy+24} proposes a comprehensive evaluation benchmark that assesses the T2V models comprehensively with multi-dimensional, human-aligned, and insight-rich properties. Although previous benchmarks present effectiveness in assessing several aspects of T2V models' capacities, most of them primarily focus on literal prompts' semantic alignment, which overlooks the integration of deeper textual and world knowledge. To address this challenge, recent work has begun to explore benchmarks for testing a model's capacity to integrate and reason according to world knowledge~\cite{mlt+24, nnz+25, zjx+25}. However, these benchmarks mainly assess T2I models. In contrast, the integration and reasoning of world knowledge in text-to-video models has not been sufficiently emphasized, which is the main motivation for our work.
\section{Benchmark}\label{sec:bench}

We present T2VWorldBench in this section, the benchmark we propose in our study. Section~\ref{sec:baseline_model} describes the baseline models. Section~\ref{sec:benchmark_prompts} provides the benchmark prompts. After that, Section~\ref{sec:evaluation_protocol} shows the evaluation protocol of our benchmark. 
 
\subsection{Baseline Models}\label{sec:baseline_model}

 In our work, we conduct a comprehensive evaluation of ten state-of-the-art text-to-video generation models released between 2024 and 2025, including commercial and open-source systems. This selection ensures that our work reflects the latest advances in T2V models, while exposing the challenges current T2V models face in integrating and reasoning about world knowledge. Details of the models are in Table~\ref{tab:models}.

To ensure consistency, videos are generated at each T2V model’s lowest available resolution, typically 720p. All videos are constrained to a 16:9 aspect ratio and limited to approximately 5 seconds. Implementation details are provided in Appendix~\ref{sec:implementation_details}.

\subsection{Benchmark Prompts}\label{sec:benchmark_prompts}

\begin{figure*}[!ht]
    \centering
    \includegraphics[width=0.9\linewidth]{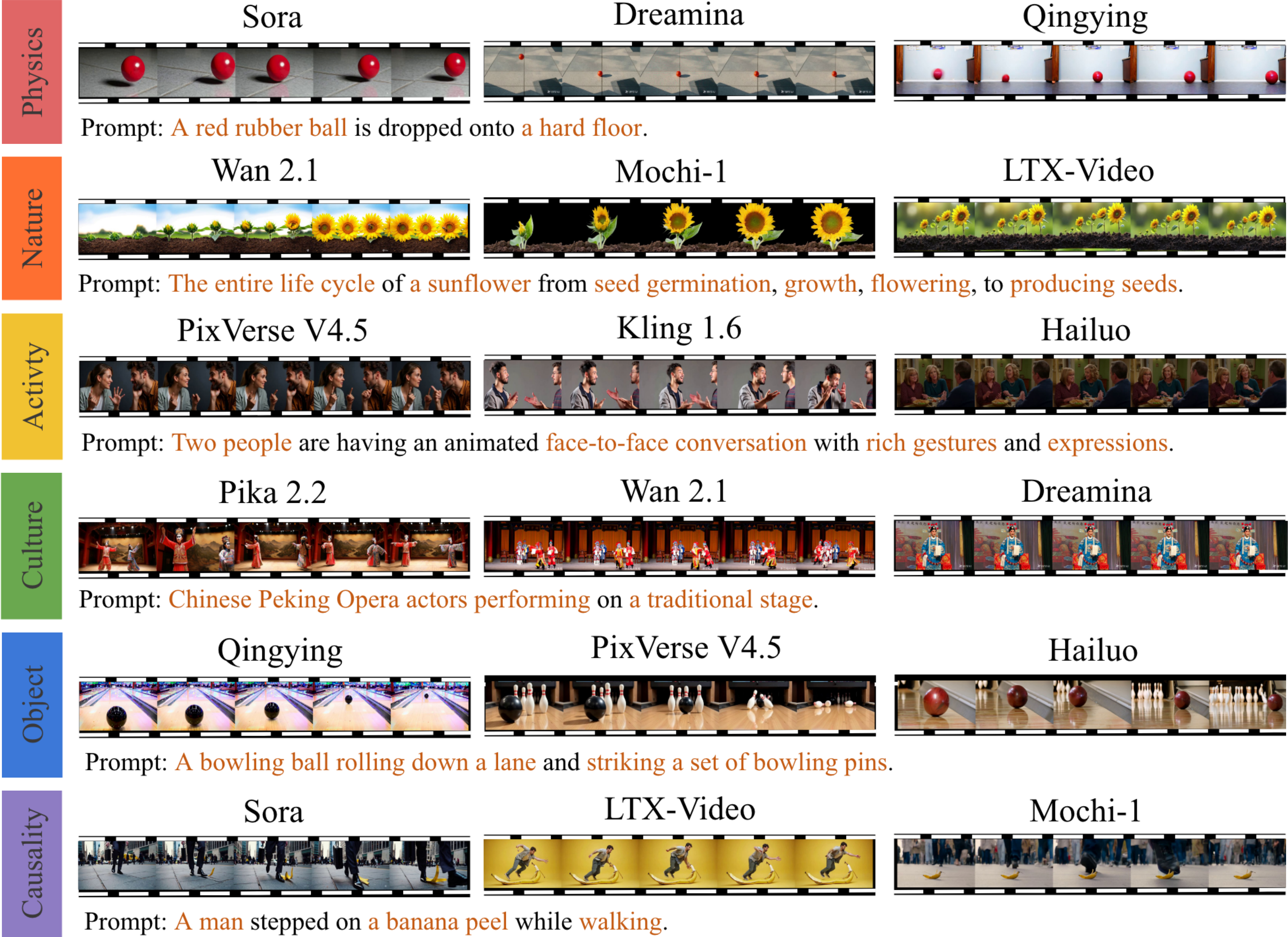}
    \caption{\textbf{Video examples from different text-to-video models are used to explain all 6 domains assessed in this benchmark.}} 
    \label{fig:benchmark_example}
\end{figure*}

 To comprehensively assess the ability of current T2V models to integrate and reason about world knowledge, the prompts need to be designed to go beyond the literal meanings. Specifically,  these prompts should challenge the T2V model's grasp of implicit knowledge, test its reasoning ability, and explore its understanding of real-world physical laws and objective facts. For instance, unlike straightforward prompts such as “A man walking”, the prompt "A man stepped on a banana peel while walking" goes beyond the literal description and requires the model to be able to infer the slippery characteristics of the banana peel, predict the resulting loss of balance, and generate a coherent video that conforms to real physical laws and causal logic. In our work, we carefully construct T2VWorldBench, which comprises 6 knowledge domains: physics, nature, activity, culture, causality, and object. Each domain includes 10 fine-grained subdomains, and each subdomain contains 20 prompts, for a total of 1,200 carefully curated prompts. Detailed composition of T2VWorldBench as shown in Figure~\ref{fig:benchmark_detail}. Figure~\ref{fig:benchmark_example} displays representative prompts and generated video results..

The key challenge in the automatic evaluation of world knowledge using VLMs is that the VLMs themselves may not possess accurate world knowledge, making their evaluations potentially unreliable. To address this issue, we provide human-authored explanations for each video, detailing the required world knowledge and reasoning chain. These explanations not only aim to clarify the implicit world knowledge and reasoning chain behind each prompt but also serve as reference points for comparing with the generation results of the T2V model. For instance, in a prompt like "A football player takes a penalty kick during a match", the corresponding explanation would specify that the video shows a soccer player on a soccer field placing the ball on the penalty spot in the penalty area, taking a few steps back, running up, and kicking the ball toward the goal while the keeper tries to pounce on it. 

{\bf Physics.} T2VWorldBench's Physics domain evaluates T2V models' ability to understand and apply basic physical principles such as gravity, motion, and force into video generation. Our goal is to investigate whether the model actually understands the physical principles in the real world, rather than just fitting an approximation to the video based on the literal meaning of the prompts over a large number of training data. For example, the T2V models should be capable of understanding the change of state from water to ice, the materials of glass and clay, and the basic laws of physics (e.g., Newton's laws) in the real world.

{\bf Nature.} The Nature domain of T2VWorldBench is designed to explore the T2V models' grasp of the natural laws and patterns that govern the living and non-living world. Our nature domain consists of 10 subdomains, such as seasons, weather, and geology. By generating videos related to natural phenomena, we attempt to investigate whether T2V models understand the causes and mechanisms of natural phenomena, rather than just generating videos with superficial meanings based on prompts. For instance, when prompted with "The entire life cycle of a sunflower from seed germination, growth, flowering, to producing seeds", an ideal text-to-video model should be able to understand and visually represent the biological processes of plant development, accurately capturing every stage from germination, stem growth to flowering, and reflecting visual realism and semantic coherence in the generated video.

{\bf Activity.} We interact with and make sense of the world is deeply related to activity, and the Activity domain of the T2VWorldBench evaluates the T2V models' ability to understand and generate coherent sequences involving activities that range from routine actions such as cooking to more structured sports. The Activity domain presents special challenges because it requires models to capture the temporal dynamics and continuous behavior of object interactions. It is not enough to recognize the surface level of actions; the T2V models must understand how the action unfolds over time and maintain consistency between video frames. For the prompt "An athlete running on the track." A reasonable generation should depict a dynamic sequence video of the athlete starting to move, maintaining a consistent running posture, and smoothly advancing along the track.

{\bf Culture.} The Culture domain focuses on evaluating the T2V models' understanding of different cultures. Significant differences in cultural customs across regions bring a great challenge for T2V models. To accurately generate culturally relevant prompts, the models need not only extensive world knowledge but also a deep understanding of the details, customs, and social aspects of each region's culture. For example, when faced with prompts such as "Celebrating Diwali in India" and "Chinese Peking Opera performance", the T2V models must recognize the related scene elements, gestures, action, and the underlying cultural meaning, in order to generate cultural topic videos that conform to objective fact.

{\bf Causality.} The Causality domain in T2VWorldBench aims to assess the T2V models' capability to understand and generate temporally and logically coherent sequences in which events are connected through causal relationships. The central challenge in this domain is whether models could recognize a single action and simulate how an action triggers a series of observable consequences that unfold over time. For a prompt like “A person knocks over a glass of water, and it spills onto the table,” the T2V model must accurately capture the continuity of the temporal sequences and reasoning the visual consequences of the process according to prompt, starting with the scene of knock over the cup and then coherently generating the scene of water spills out.

{\bf Object.} Finally, T2VWorldBench features an Object domain that evaluates whether T2V models can go beyond basic object recognition to demonstrate a deeper understanding of visual and physical object properties, including space, appearance, quantity, size, perspective, structure, shape, interaction motion, and function. For example, when given the prompt "A person using a pair of scissors to cut a piece of paper", a text-to-video model that understands world knowledge should not only recognize the individual objects (person, scissors, and paper) but should also understand their formal characteristics and functional interactions. This includes correctly depicting the way the person holds the scissors, the mechanical movement of the scissors, the deformation and fracture of the paper during the cutting process, and maintaining temporal coherence throughout the cutting process, all of which together represent the model's capability to integrate world knowledge to model the properties of the objects, uses, and the logic of their interactions.

\subsection{Evaluation Protocol}\label{sec:evaluation_protocol}

Previous benchmark evaluations have scored the model generation results with a single metric~\cite{uvk+19}, which serves as an effective and intuitive method for assessing the generation results of the T2V models. However, such a single assessment metric oversimplifies the assessment of the T2V models' capabilities and may fail to provide a comprehensive presentation of the T2V models' specific capabilities in several areas. In our work, we introduce four evaluation dimensions: video quality, video realism, video relevance, and video consistency. 

To ensure the robustness and reliability of the assessment, we used an assessment strategy combining human and automatic assessment. We provide a detailed introduction to the human and automatic evaluation protocols as follows:

{\bf Human Evaluation.} 

To better align with human preference and get excellent evaluation results, we incorporate two independent human annotators in our evaluation process, who have expertise in AI and have normal vision capability (not disabled). These human annotators independently evaluate the generated result from multiple dimensions, including video quality, video realism, video relevance, and video consistency. For each evaluation dimension, we provide five levels of ratings, which are subsequently normalized to scores in the range [0, 1]: 

\begin{itemize}
    \item {\bf Level 1} (scores 0.2): Poor, the video presents very low video quality, with severe motion artifacts such as blurring and dragging, lacking visual authenticity, and is not generated according to prompts. The video frames are clearly inconsistent, making the content incoherent and difficult to follow.
    \item {\bf Level 2} (scores 0.4): Fair, the video has obvious quality issues, such as blurred focus and unclear screen. The visual effect is somewhat artificial and lacks convincing realism. Although the generated video is related to the prompt, the connection is weak. The transition between frames is not continuous enough.
    \item {\bf Level 3} (scores 0.6): Acceptable, the video achieves a basic level of quality, with generally recognizable content and acceptable visual defects, and the video roughly captures the primary meaning of the prompt, although some details may be incomplete. The video's realism is acceptable, broadly consistent with artificial preferences. Video consistency is generally acceptable, but there are still some disjointed scenes.
    \item {\bf Level 4} (scores 0.8): Good, the video delivers clear screens, there are only a few minor visual flaws, and it aligns well with the prompt, generating video that matches the world knowledge. The video maintains a credible level of realism, the progress across frames is mostly smooth, creating a coherent and engaging video.
    \item {\bf Level 5} (scores 1.0): Excellent, the video stands out with vivid, high-quality visual effects and a high attention to detail. It not only matches the prompt accurately, but also conveys its implicit world knowledge through realistic and expressive content. Each frame naturally transitions to the next frame, reflecting a high degree of consistency in time.
\end{itemize}

{\bf Automatic Evaluation.} 

To enable scalable and efficient evaluation of text-to-video models and learn automatic assessment techniques from previous excellent work~\cite{hhy+24, shl+25}, we introduce the current well-performing multimodal LLaVA1.6-34B~\cite{llll24} model for automated evaluation to fully utilize its advantages in visual and linguistic fusion comprehension to achieve accurate analysis and scoring of the generated video content, and to enhance automation and consistency of the evaluation process.

To guide the automated evaluation, we constructed fine-grained reference explanations for each prompt. These explanations detail what an ideal video, which is based on world knowledge and relevant schemas, should be like. By grounding the assessment in explicit, knowledge-based expectations, we ensure that the assessment captures not only surface visual features but also semantic consistency and relevance to the prompt. Importantly, the evaluation metrics used in the automatic setting are aligned with those used in human evaluation to ensure comparability and reliability across both evaluation methods.

\begin{figure*}[!ht]
    \centering
    \includegraphics[width=0.9\linewidth]{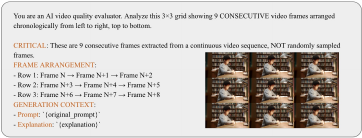}
    \caption{{\bf Template for Base Prompt.} The Base Prompt establishes the foundational context for all evaluation tasks. It instructs the AI evaluator on its role, defines the input structure as a 3 $\times$ 3 grid of nine consecutive video frames, visually exemplified on the right, and provides the original prompt and explanation. This prompt serves as the common groundwork upon which all specific dimensional evaluations are built.} 
    \label{fig:base_prompt}
\end{figure*}

Our general way of using VLM begins with meta-prompts designed to guide the model in structured evaluation of the generated video. Concretely, we adopt a two-stage prompting strategy consisting of a base prompt and a set of task-specific prompts corresponding to four key evaluation metrics: video quality, realism, relevance, and consistency. First, to evaluate an entire video, we segment it into grids, where each grid consists of 9 consecutive frames arranged in a 3 $\times$ 3 layout. The VLM then sequentially evaluates every grid from the video, and the minimum score from among all grids is adopted as the final score for the video. For each grid, the evaluation begins with a base prompt that establishes the context. This prompt includes the original text-to-video prompt and a fine-grained explanation derived from world knowledge, which together define the intended semantics of the video. See Figure~\ref{fig:base_prompt} for reference.

At the second step of the automatic evaluation, we provide the evaluation model with detailed metric-specific prompts corresponding to the four key evaluation dimensions (video quality, realism, relevance, and consistency). Each prompt is carefully designed to direct the model to focus on specific aspects of the video, thus making the automatic evaluation more targeted and interpretable. The automatic evaluation prompts for each dimension are detailed below: 

\begin{itemize}
    \item {\bf Quality}: This evaluation prompt assesses the technical fidelity of the generated video. A high-quality video that conforms to human preferences should present clear visuals with minimal distortion or noise. The evaluation model is instructed to focus on low-level visual attributes such as resolution, sharpness, clarity, and the presence of rendering artifacts (e.g., burrs, blurring, or broken geometry). 
    \item {\bf Realism}: This aspect focuses on the visual believability of the generated content. The focus is on whether the video looks realistic and naturally occurring, taking into account factors such as realistic object textures, lighting and shadows, physical interactions, and adherence to common-sense physics. The goal is to ensure that the generated scenes and objects do not appear artificial, absurd, or physically unbelievable.
    \item {\bf Relevance}: This dimension focuses on the alignment between the input prompt and the content of the generated video, along with fine-grained explanations. The focus of evaluation is whether the key entities, actions, and scenes inferred based on world knowledge prompts are correctly presented in the generated video. The video is expected to accurately and comprehensively reflect the intended semantics and fine-grained details conveyed by the prompt, and can perfectly showcase the relevant background knowledge in the prompt.
    \item {\bf Consistency}: This evaluation prompt is designed to assess the temporal consistency of sequential frames in a video, analyzing whether the objects in the generated video maintain their state, position, appearance, and motion continuity over time. A high degree of consistency is essential to convey believable video meaning. 
\end{itemize}

The template for the Stage 2 evaluation prompt is provided in Appendix~\ref{sec:evaluate_prompt}.

To obtain a robust and balanced final evaluation of the performance of the T2V models, we adopted a hybrid scoring protocol that combines human evaluation and automatic evaluation. For each evaluation dimension, namely quality, realism, and relevance, we first aggregate human annotations by calculating the average score assigned by the raters. Then, the artificially obtained scores are fused with the corresponding scores generated by the automatic evaluation framework to ensure semantic sensitivity and consistency. Each dimension’s final score is derived by taking the average of manual and automatic assessments. Subsequently, The comprehensive evaluation score for generated videos is obtained by taking the average of the four dimensions:

\begin{align}\label{eq:score}
    S_{overall} = \frac{1}{4} \sum_{d \in D} S_d,
\end{align}

where $D$ = \{Quality, Realims, Relevance, Consistency\}. Our evaluation strategy implements a comprehensive evaluation process that combines subjective human insight with scalable and repeatable automated assessment.

\section{Experiments}\label{sec:experiments}

We show primary experimental results of the T2VWorldBench in this section. We analyze and discuss the insights derived from the comprehensive results in Section~\ref{sec:overall_result}. We provide qualitative comparisons between correct and incorrect generations in Section~\ref{sec:quantitative_study}. Finally, we analyze the variance among human annotators in Section~\ref{sec:annotator_analysis}.

\subsection{Overall World Knowledge Result}\label{sec:overall_result}

\begin{table*}[!ht]
\centering
\resizebox{1.0\linewidth}{!}{
\begin{tabular}{|c|cccccc|c|}
\toprule
\textbf{Model} & \textbf{Physics} & \textbf{Nature} & \textbf{Activity} & \textbf{Culture} & \textbf{Causality} & \textbf{Object} & \textbf{Avg.} \\
\midrule
Wan2.1        & 0.70 & 0.68 & 0.71 & 0.67 & 0.62 & 0.72 & 0.68 \\
LTX Video     & 0.65 & 0.68 & 0.73 & 0.66 & 0.65 & 0.68 & 0.68 \\
Kling1.6         & 0.66 & 0.68 & 0.74 & 0.69 & 0.62 & 0.66 & 0.67 \\
Dreamina      & 0.63 & 0.68 & 0.69 & 0.68 & 0.63 & 0.69 & 0.67 \\
Mochi-1       & 0.63 & 0.72 & 0.68 & 0.63 & 0.62 & 0.68 & 0.66 \\
Sora          & 0.64 & 0.69 & 0.67 & 0.67 & 0.57 & 0.64 & 0.65 \\
Hailuo        & 0.60 & 0.62 & 0.68 & 0.65 & 0.58 & 0.65 & 0.63 \\
PixVerse V4.5   & 0.59 & 0.64 & 0.66 & 0.62 & 0.58 & 0.68 & 0.63 \\
Qingying      & 0.57 & 0.57 & 0.63 & 0.64 & 0.56 & 0.68 & 0.61 \\
Pika2.2       & 0.61 & 0.73 & 0.60 & 0.57 & 0.56 & 0.56 & 0.60 \\
\bottomrule
\end{tabular}
}
\caption{\bf Model performance across six dimensions and overall average.}
\label{tab:dimension_scores}
\end{table*}

\begin{figure}[!ht]
    \centering
    \includegraphics[width=0.8\linewidth]{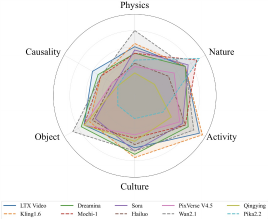}
    \caption{\bf Radar Plot of Model Performance Across 6 Evaluation Dimensions.}
    \label{fig:radar}
\end{figure}

Scores for each evaluation dimension are derived following the evaluation protocol illustrated in Section~\ref{sec:evaluation_protocol}. The resulting outcomes are present in Table~\ref{tab:dimension_scores} and Figure~\ref{fig:radar}.

As shown in Table~\ref{tab:dimension_scores}, current text-to-video models still face significant challenges in generating videos based on prompts that require the integration of world knowledge. Even the best-performing models in our benchmark, such as Wan2.1 and LTX Video, show only moderate performance with an average score of about 0.68, indicating that there is much room for improvement. This brings us to the following insight:

\begin{observation}
    The overall score of  SOTA text-to-video models is not ideal, and the most advanced T2V models are still far from mastering the ability of world knowledge-intensive generation, highlighting their huge gap in combining world knowledge reasoning.
\end{observation}

Across all text-to-video models in our benchmark, the relatively strong performance in evaluation domains such as activity and object suggests that the current text-to-video models are better at capturing surface-level actions and visual objects. In contrast, scores for evaluation dimensions such as causality and culture are consistently low, highlighting the difficulty of modeling abstract reasoning and cultural scenes. For instance, both Sora and Pika2.2 are below 0.60 on the causality dimension, emphasizing the limitations of dealing with complex events where multiple factors interact. Furthermore, while some dedicated text-to-video models, such as LTX Video and Wan2.1, performed competitively on most evaluation dimensions, other models exhibit significant variability depending on the different evaluation dimensions. This shows that current text-to-video models typically excel in narrowly defined capabilities but lack generalized robustness across different types of prompts that require a nuanced understanding of world knowledge. These findings point to the following observation:

\begin{observation}
    While current text-to-video models perform reasonably well at the visual level, they struggle with world knowledge reasoning and exhibit limited generalization capabilities.
\end{observation}

\subsection{Quantitative Study}\label{sec:quantitative_study}

To better understand how text-to-video models perform when faced with world knowledge prompts, we conduct a qualitative study based on our proposed benchmark, T2VWorldBench. Unlike existing benchmarks that tend to focus on generic scenarios, such as object motion or visually simple activities. T2VWorldBench challenges models to create videos based on real-world understanding, cultural, and contextual knowledge. For instance, a typical prompt such as “the type of cutlery commonly used by East Asians when eating”, which requires the model to correctly link cultural knowledge (e.g., chopstick use) to visual generation. This shift toward world knowledge video synthesis allows us to more challengingly and meaningfully assess whether models can go beyond literal relevance to incorporate general, cultural, and natural knowledge into coherent visual outputs.

Furthermore, we analyze the correct and incorrect generation results produced by different T2V models under the same prompts in our benchmark, highlighting the advantages and limitations of current T2V models in understanding and reasoning world knowledge, as shown in Figure~\ref{fig:true} and Figure~\ref{fig:false} in Appendix~\ref{sec:video_examples}.

For the prompt “A red rubber ball falls onto a hard floor”, Mochi~\cite{mochi1} demonstrates an accurate understanding by generating a coherent video depicting the expected motion. In contrast, PixVerse~\cite{pixVerse} fails to capture the underlying dynamics and instead generates a red ball rolling on a floor, missing the motion and interaction implied by the prompt. Similarly, for prompts requiring cultural or natural world knowledge, the differences became more pronounced. When prompted with “the US President's workplace”, Qingying~\cite{qingying} generates a random white palatial structure loosely associated with the concept, while Sora~\cite{sora} successfully identifies and visualizes the White House, demonstrating a higher level of knowledge base and ability to reason with world knowledge. In another example, the prompt “the most common spiky plant in the desert” reveals Kling's~\cite{kling} ability to associate the description with a cactus, demonstrating correct visual semantic alignment, while Dreamina~\cite{dreamina} depicts a giant, spiky, abstract green plant that lacks real-world rationality. A similar contrast can be seen in the physical causality scene "A man stepped on a banana peel while walking": Wan~\cite{wan2.1} captures the full causal sequence through a realistic slip and fall, while Hailuo~\cite{hailuo} only shows a person passing by a banana peel, failing to capture the implied causa l sequence and the expected outcome embedded in the prompt. This brings us to the following insight:

\begin{observation}
    For current text-to-video models, even if they can understand the semantics when dealing with prompts that contain world knowledge, they are often biased in the generation stage, outputting videos that do not conform to reality or logic, exposing a significant gap between understanding and generation.
\end{observation}

\subsection{Human Annotator Variance Analysis}\label{sec:annotator_analysis}

For each evaluation dimension, we take the average of the two annotators’ scores to produce a more stable and representative assessment. To assess the consistency and reliability between the two annotators, we through Eq.~\eqref{eq:pearson} computing the Pearson correlation coefficient (r) over their scores across all evaluation dimensions. 

We define two annotators’ scores for each evaluation metric as $X = \{X_1, X_2, \ldots, X_n\}$ and $Y = \{Y_1, Y_2, \ldots, Y_n\}$, where $n$ is the number of samples. The Pearson correlation coefficient $r$ between their scores is calculated as follows:

\begin{align}\label{eq:pearson}
    r = \frac{\sum_{i=1}^{n} (X_i - \bar{X})(Y_i - \bar{Y})}
         {\sqrt{\sum_{i=1}^{n} (X_i - \bar{X})^2} \cdot 
          \sqrt{\sum_{i=1}^{n} (Y_i - \bar{Y})^2}}.
\end{align}

\begin{table}[!ht]
\centering
\begin{tabular}{|c|c|c|}
\hline
\textbf{Metric} & \textbf{Pearson (r)} & \textbf{Agreement Level} \\
\hline
Quality     & 0.623 & Moderate agreement \\
Realism     & 0.617 & Moderate agreement \\
Relevance   & 0.728 & Substantial agreement \\
Consistency & 0.758 & Substantial agreement \\
\hline
\end{tabular}
\caption{{\bf Pearson correlation coefficients.} This table reports the Pearson correlation coefficients between two human annotators for each evaluation metric, along with the corresponding qualitative interpretation of agreement levels.}
\label{tab:pearson_interp}
\end{table}
\section{Conclusion}\label{sec:conclusion}

In our study, we propose {\bf T2VWorldBench}, an innovative benchmark created to systematically evaluate the ability of text-to-video models to understand and integrate world knowledge, comprising 1,200 prompts from 6 dimensions and 60 sub-dimensions. Our evaluation of 10 text-to-video models, spanning both commercial and open-source models, reveals that there are still significant shortcomings in their effective utilization of world knowledge in the generation process. Even the text-to-video models with the best overall performance to date struggle to demonstrate excellent video generation performance in complex and knowledge-dependent reasoning scenarios. We hope that our work can provide a reference for future research and stimulate further exploration and improvement in enhancing the world knowledge understanding and synthesis ability of text-to-video models.


\newpage
\onecolumn
\appendix

\begin{center}
    \textbf{\LARGE Appendix }
\end{center}


{\bf Roadmap.} 
Section~\ref{sec:implementation_details} introduces the ten baseline text-to-video models' implementation details. In Section~\ref{sec:evaluate_prompt}, we present the evaluation prompt template. In Section~\ref{sec:video_examples}, we display a range of video examples.

\section{Implementation Details} \label{sec:implementation_details}

This section provides further details on the 10 baseline text-to-video models, as listed below:

\begin{itemize}
    \item {\bf Sora}~\cite{sora}: Developed by the OpenAI team in 2024, Sora is a closed-source generative model. The model can generate 30 FPS videos with selectable durations of 5, 10, 15, or 20 seconds. It supports a range of output formats, including resolution from 480p to 1080p and multiple aspect ratios (1:1, 16:9, and 9:16). Sora provides style presets and can produce four distinct video variants from a single prompt. Additionally, a ``relaxed mode" is available with a processing latency of approximately 30 seconds per video.

    \item {\bf Dreamina Video 3.0}~\cite{dreamina}: Dreamina Video 3.0 is a closed-source generator released by the Bytedance team in 2024, supporting both 5- and 10-second videos. It supports a wide range of aspect ratios (16:9, 21:9, 4:3, 1:1, 3:4, 9:16) and utilizes DeepSeek-R1~\cite{gyz+25} for prompt enhancement. 

    \item {\bf Qingying}~\cite{qingying}: Qingying is the commercial implementation of Zhipu's open-source CogVideo models~\cite{hdz+23,ytz+24}. It generates 5-second videos at 30/60 FPS across five aspect ratios of 1:1, 9:16, 16:9, 4:3, and 3:4. Qingying supports two modes: Quality and Fast. Additionally, it provides fine-grained control over video style, emotional atmosphere, and camera movement, alongside support for AI-generated audio and visual effects.

    \item {\bf Wan2.1 Plus}~\cite{wan2.1}: Wan2.1 Plus is an open-source generative model~\cite{wan_open} released by Alibaba Group in 2025, supporting multiple aspect ratios (1:1, 3:4, 4:3, 9:16, 16:9). It provides additional features such as ``Inspiration Mode" and ``Sound Effects". 

    \item {\bf Mochi-1}~\cite{mochi1}: Released by Genmo in 2024, Mochi-1 is an open-source model. Its standard output consists of 5-second, 24 FPS video at 480p resolution with a 16:9 aspect ratio. Mochi-1 supports a seed function for reproducibility and includes a feature for random prompt suggestions. It can generate two videos simultaneously, with an approximate processing time of 3 minutes per video.

    \item {\bf LTX Video}~\cite{hcb+24}: Developed by Lightricks in 2024, LTX Video is an open-source model. It generates 5-second, 24 FPS videos at 512p resolution, supporting 16:9, 1:1, and 9:16 aspect ratios. LTX Video enables fine-grained control over the location, shot type, references, and style, and even supports voiceover integration.

    \item {\bf PixVerse V4.5}~\cite{pixVerse}: PixVerse V4.5 is a closed-source model from AISphere, released in 2025. It generates videos with a duration of either 5 or 8 seconds. PixVerse V4.5 supports multiple resolutions including 360p, 540p, 720p, and 1080p, and offers five aspect ratios: 16:9, 4:3, 1:1, 3:4, and 9:16.

    \item {\bf Kling 1.6}~\cite{kling}: Released by Kuaishou in 2024, Kling 1.6 is a closed-source generative model. It generates video outputs of 5 or 10 seconds in duration, supporting 16:9, 1:1, and 9:16 aspect ratios. It features two generation modes: a standard mode and a restricted high-quality mode. Kling supports advanced prompting functionalities, including negative prompts, fixed seeds for reproducibility, a prompt dictionary, and AI-assisted prompt suggestions. For generations, Kling can create 4 videos simultaneously from a single prompt. The processing time is approximately 4 minutes per video, with a maximum batch size of 5 videos.
   
    \item {\bf Hailuo 01-Director}~\cite{hailuo}: Hailuo 01-Director is a closed-source model released by Minimax in 2025 for text-to-video generation. Its standard output is a 6-second, 24 FPS video at 720p resolution, typically with a default aspect ratio of 16:9.

    \item {\bf Pika 2.2}~\cite{pika2.2}: Pika2.2 is a closed-source generative model from Pika Labs, released in 2025. It provides  Pikawaps, Pikaaddition, Pikaaffects, Pikaframes, and Pikascenes. Pika 2.2 generates 5- or 10-second videos at resolutions of 720p or 1080p, supporting a wide range of aspect ratios (16:9, 9:16, 1:1, 4:5, 4:3, 5:2). For generation control, it supports both negative prompts and seed inputs. Pika 2.2 can produce 4 videos simultaneously, with each taking approximately 30 seconds to process.
\end{itemize}

\section{Evaluation Prompt}\label{sec:evaluate_prompt}

\begin{figure}[!ht]
    \centering
    \includegraphics[width=1.0\linewidth]{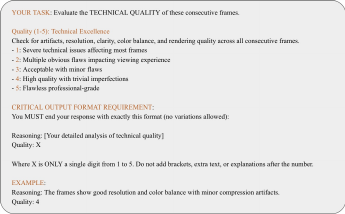}
    \caption{{\bf Template for Quality Prompt.} Quality Prompt is designed specifically for evaluating the video quality. It directs the AI evaluator to assess objective attributes such as artifacts, resolution, clarity, and color balance, using a 1-to-5 score to quantify the video quality from severely flawed to professional-grade.} 
    \label{fig:quality_prompt}
\end{figure}

\begin{figure}[!ht]
    \centering
    \includegraphics[width=1.0\linewidth]{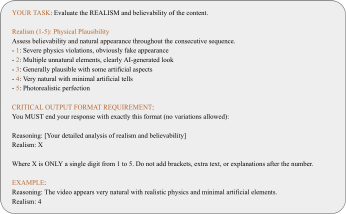}
    \caption{{\bf Template for Realism Prompt.} Realism Prompt guides the evaluation of the video realism. The assessment is based on physical plausibility, instructing the evaluator to identify any physics violations, unnatural elements, or other artificial tell. The 1-to-5 score quantifies how closely the content approximates photorealistic perfection.} 
    \label{fig:realism_prompt}
\end{figure}

\begin{figure}[!ht]
    \centering
    \includegraphics[width=1.0\linewidth]{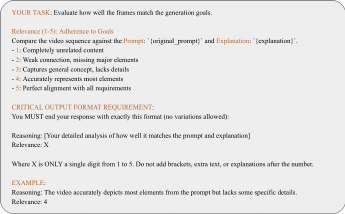}
    \caption{{\bf Template for Relevance Prompt.} Relevance Prompt focuses on evaluating video relevance to the user's prompt and explanation. It instructs the AI evaluator to compare the visual output against the provided original prompt and explanation, scoring the alignment on a 1-to-5 scale based on how accurately the generated content captures the required elements and world knowledge.}
    \label{fig:relevance_prompt}
\end{figure}

\begin{figure}[!ht]
    \centering
    \includegraphics[width=1.0\linewidth]{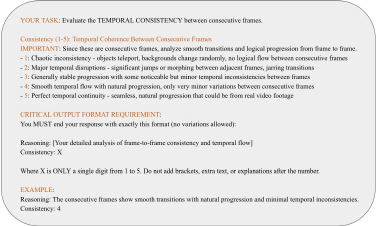}
    \caption{{\bf Template for Consistency Prompt.} Consistency Prompt is used to evaluate the video consistency. The core task is to analyze the coherence between consecutive frames, focusing on the smoothness of transitions and the logical progression of objects and actions. The 1-to-5 score measures the video's temporal flow, from chaotic and disjointed to seamless and natural.} 
    \label{fig:consistency_prompt}
\end{figure}

We employ a prompt-based framework to conduct a comprehensive evaluation of AI-generated videos using the LLaVA model. This methodology involves combining a base prompt (see Figure~\ref{fig:base_prompt}) with a specialized prompt tailored to a specific evaluation dimension. For instance, evaluating video quality is achieved by pairing the Base Prompt with the Quality Prompt (see Figure~\ref{fig:quality_prompt}). Similarly, video realism is evaluated using the Base Prompt plus the Realism Prompt (see Figure~\ref{fig:realism_prompt}). The same approach is applied to evaluate video relevance using the Relevance Prompt (see Figure~\ref{fig:relevance_prompt}) and consistency using the Consistency Prompt (see Figure~\ref{fig:consistency_prompt}). This ensures that each evaluation is grounded in a consistent context while allowing for a focused, independent score for each distinct attribute of the video.

\section{Video Examples}\label{sec:video_examples}

\begin{figure}[!ht]
    \centering
    \includegraphics[width=1.0\linewidth]{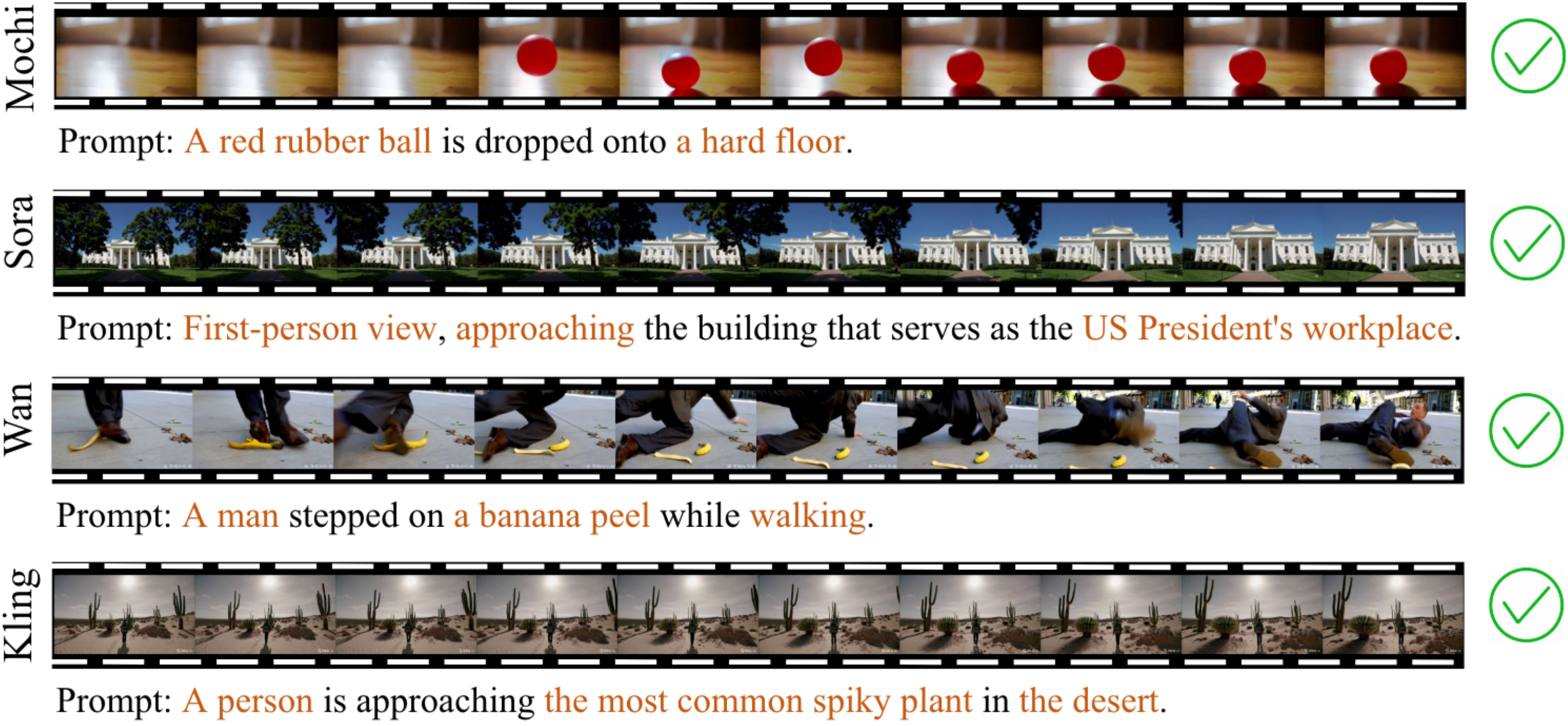}
    \caption{Examples of successfully understanding world knowledge.} 
    \label{fig:true}
\end{figure}

\begin{figure}[!ht]
    \centering
    \includegraphics[width=1.0\linewidth]{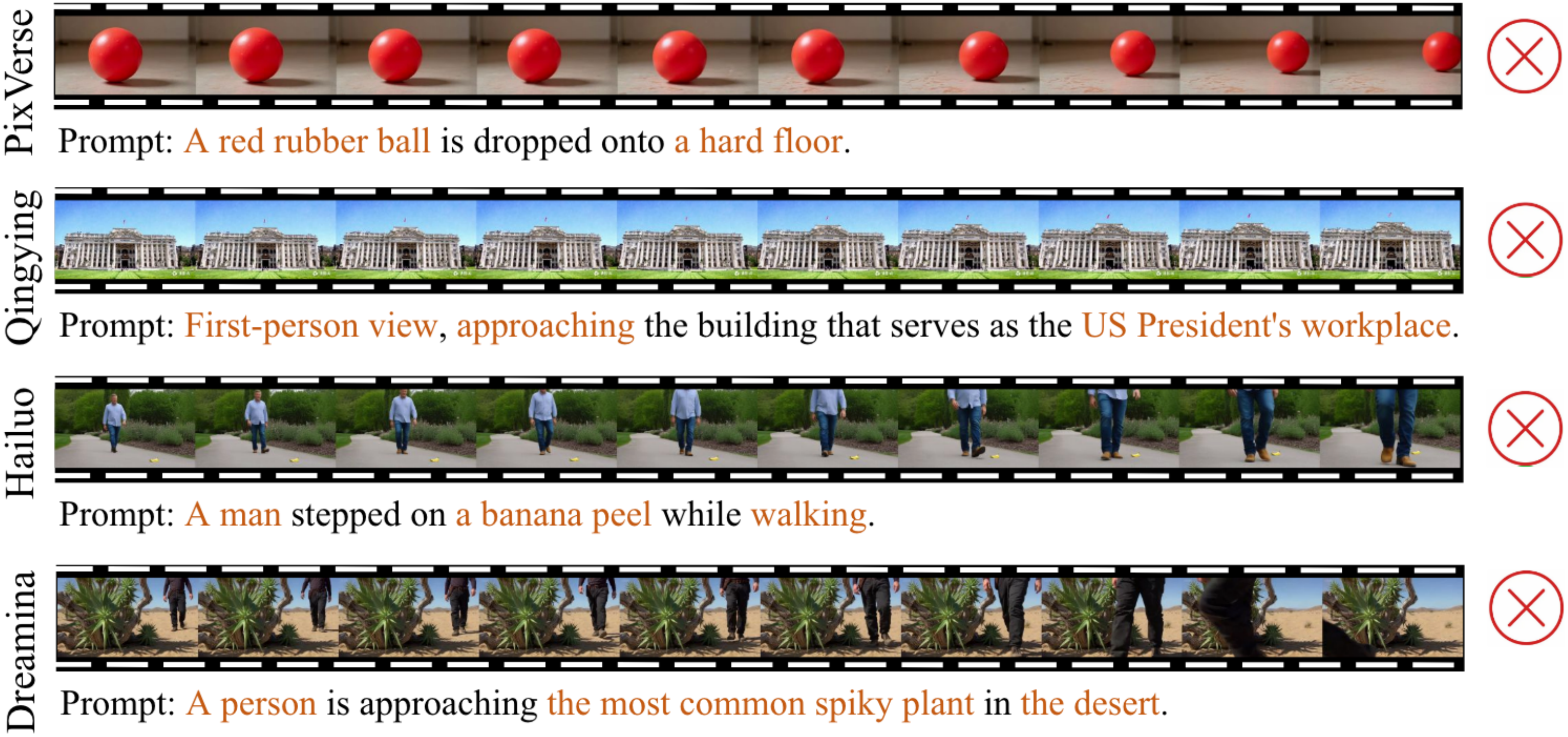}
    \caption{Examples of failures in understanding world knowledge.} 
    \label{fig:false}
\end{figure}

\begin{figure}[!ht]
    \centering
    \includegraphics[width=1.0\linewidth]{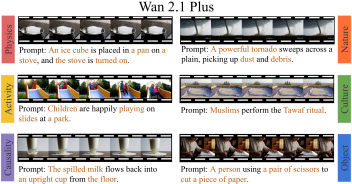}
    \caption{Video generation of Wan 2.1 Plus.} 
    \label{fig:wan}
\end{figure}

\begin{figure}[!ht]
    \centering
    \includegraphics[width=1.0\linewidth]{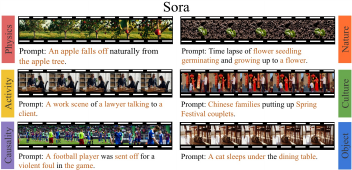}
    \caption{Video generation of Sora.}
    \label{fig:sora}
\end{figure}

\begin{figure}[!ht]
    \centering
    \includegraphics[width=1.0\linewidth]{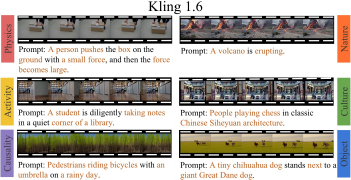}
    \caption{Video generation of Kling 1.6.} 
    \label{fig:kling}
\end{figure}

\begin{figure}[!ht]
    \centering
    \includegraphics[width=1.0\linewidth]{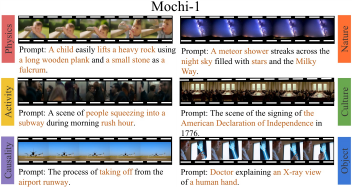}
    \caption{Video generation of Mochi-1.} 
    \label{fig:mochi}
\end{figure}

\begin{figure}[!ht]
    \centering
    \includegraphics[width=1.0\linewidth]{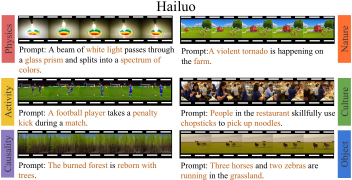}
    \caption{Video generation of Hailuo.} 
    \label{fig:hailuo}
\end{figure}

\begin{figure}[!ht]
    \centering
    \includegraphics[width=1.0\linewidth]{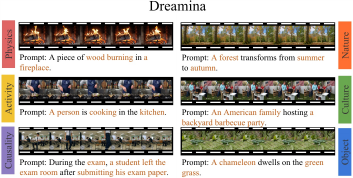}
    \caption{Video generation of Dreamina.} 
    \label{fig:dreamina}
\end{figure}

\begin{figure}[!ht]
    \centering
    \includegraphics[width=1.0\linewidth]{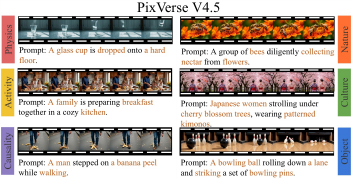}
    \caption{Video generation of PixVerse V4.5.} 
    \label{fig:pixverse}
\end{figure}

\begin{figure}[!ht]
    \centering
    \includegraphics[width=1.0\linewidth]{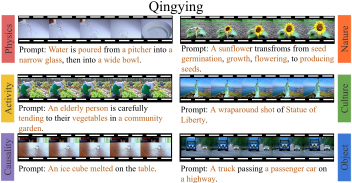}
    \caption{Video generation of Qingying.} 
    \label{fig:qingying}
\end{figure}

\begin{figure}[!ht]
    \centering
    \includegraphics[width=1.0\linewidth]{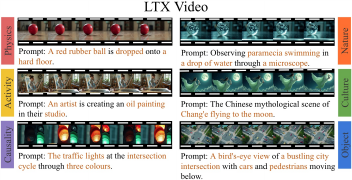}
    \caption{Video generation of LTX Video.} 
    \label{fig:ltxvideo}
\end{figure}

\begin{figure}[!ht]
    \centering
    \includegraphics[width=1.0\linewidth]{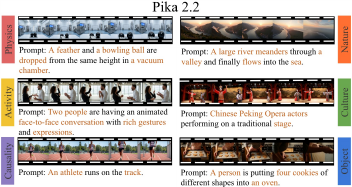}
    \caption{Video generation of Pika 2.2.} 
    \label{fig:pika}
\end{figure}

In this Section, we provide extensive examples of videos generated by our proposed benchmark prompts. Figure~\ref{fig:true} and Figure~\ref{fig:false} present the results of our quality study. Figure~\ref{fig:wan}-\ref{fig:pika} shows the generation result of each text-to-video model in our benchmark, where five representative frames from each video are extracted and arranged sequentially to form a visual strip. These presented instances are consistent with the experimental setting discussed in Section~\ref{sec:experiments}


\newpage
\clearpage
\ifdefined\isarxiv
\bibliographystyle{alpha}
\bibliography{ref}
\else
{\small
\bibliographystyle{ieeenat_fullname}
\bibliography{ref}
}
\fi

\end{document}